\documentclass[10pt,twocolumn,letterpaper]{article}

\usepackage{iccv}
\usepackage{times}
\usepackage{epsfig}
\usepackage{graphicx}
\usepackage{amsmath}
\usepackage{relsize}
\usepackage{amssymb}
\usepackage{caption}
\usepackage{subcaption}

\usepackage[breaklinks=true,bookmarks=false]{hyperref}

\iccvfinalcopy 

\def\iccvPaperID{8737} 

\usepackage[font=small,labelfont=bf]{caption}
\ificcvfinal\fi

\begin{document}

\newif\ifdraft
\draftfalse

\ifdraft
    \newcommand{\yossi}[1]{{\color{green}\textbf{Yossi:} #1}}
    \newcommand{\assaf}[1]{{\color{blue}\textbf{Assaf:} #1}}
    \newcommand{\amil}[1]{{\color{red}\textbf{Amil:} #1}}
    
    \newcommand{\sasha}[1]{{\color{yellow}\textbf{Sasha:} #1}}

\else
    \newcommand{\yossi}[1]{}
    \newcommand{\assaf}[1]{}
    \newcommand{\amil}[1]{}
    \newcommand{\sasha}[1]{}

\fi

\title{
     Rosetta Neurons: Mining the Common Units in a Model Zoo

}

\author{%
  Amil Dravid$^{*}$ \\
  Northwestern \\
   \and
  Yossi Gandelsman$^{*}$ \\
  UC Berkeley \\
  \and 
  Alexei A. Efros \\
  UC Berkeley\\
   \and
  Assaf Shocher \\
  UC Berkeley, Google 
}

\twocolumn[{%
\renewcommand\twocolumn[1][]{#1}%
\maketitle
\begin{center}
    \centering
    \vspace{-1.cm}
    \captionsetup{type=figure}
    \includegraphics[width=0.97\textwidth]{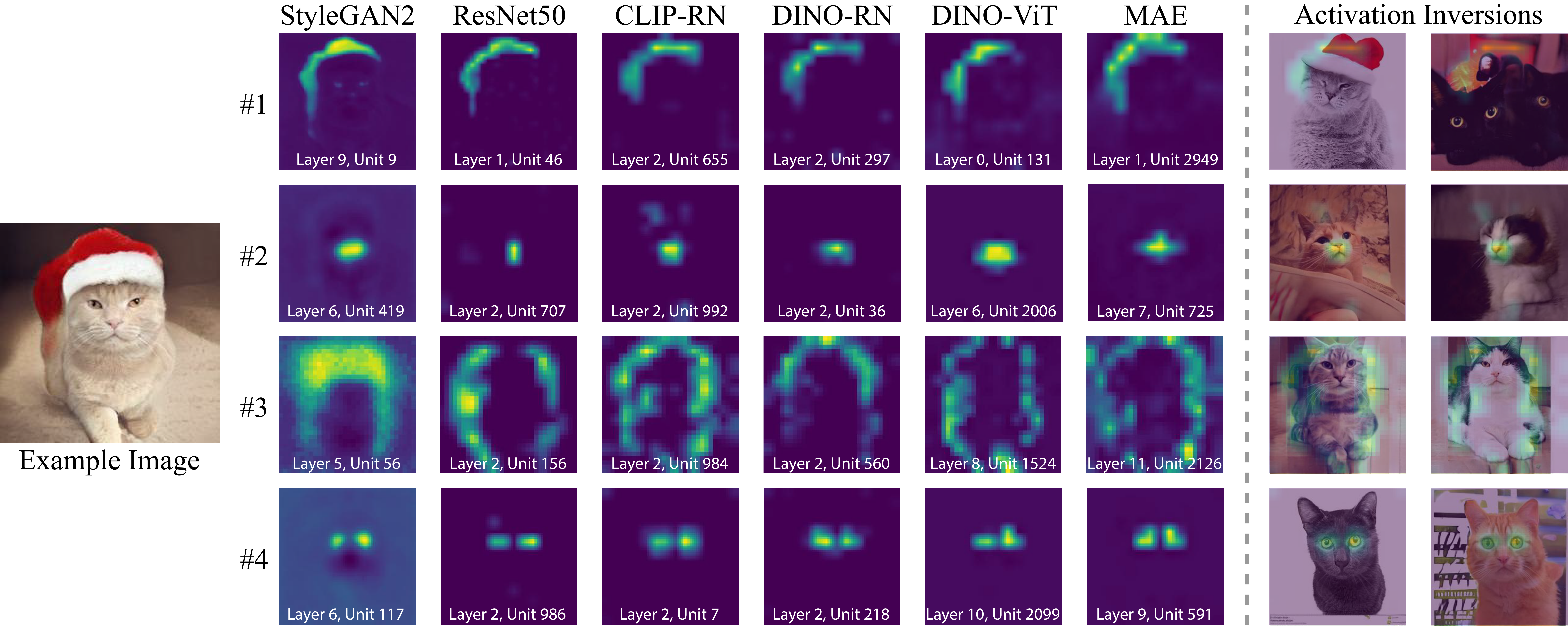}
     \vspace*{-0.3cm}
    \captionof{figure}{\textbf{Mining for ``Rosetta Neurons.''} Our findings demonstrate the existence of matching neurons across different models that express a shared concept (such as object contours, object parts, and colors). These concepts emerge without any supervision or manual annotations. We visualize the concepts with heatmaps and a novel inversion technique (two right columns).
    }
    \label{fig:teaser}
\end{center}%
}]

\ificcvfinal\thispagestyle{empty}\fi

\newcommand\blfootnote[1]{%
  \begingroup
  \renewcommand\thefootnote{}\relax\footnotetext{#1}%
  \addtocounter{footnote}{0}%
  \endgroup
}

\begin{abstract}
\vspace{-0.2cm}
\let\thefootnote\relax\footnotetext{* Equal contribution.}
\addtocounter{footnote}{0}\let\thefootnote\svthefootnote
Do different neural networks, trained for various vision tasks, share some common representations?
   In this paper, we demonstrate the existence of common features we call {\em ``Rosetta Neurons"} across a range of models with different architectures, different tasks (generative and discriminative), and different types of supervision (class-supervised, text-supervised, self-supervised). We present an algorithm for mining a dictionary of Rosetta Neurons across several popular vision models:
      Class Supervised-ResNet50, DINO-ResNet50, DINO-ViT, MAE, CLIP-ResNet50, BigGAN, StyleGAN-2, StyleGAN-XL.
   Our findings suggest that certain visual concepts and structures are inherently embedded in the natural world and can be learned by different models regardless of the specific task or architecture, and without the use of semantic labels. We can visualize shared concepts directly due to generative models included in our analysis. The Rosetta Neurons facilitate model-to-model translation enabling various inversion-based manipulations, including cross-class alignments, shifting, zooming, and more, without the need for specialized training.
\end{abstract}


\begin{figure*}
\vspace*{-0.2cm}
  \centering
  \hspace*{-0.2cm} \includegraphics[width=\textwidth]{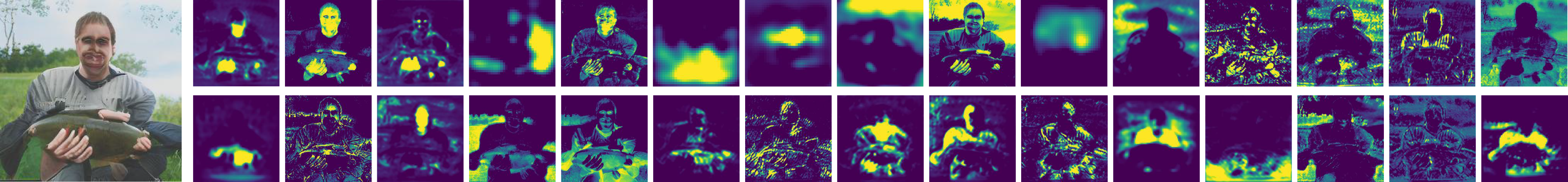}
\caption{\label{fig:all_neurons}
\textbf{Visualization of all the concepts for one class.} An example of the set of all concepts emerging for ImageNet ``Tench'' class by matching the five discriminative models from Table \ref{tab:models} and clustering within StyleGAN-XL. GAN heatmaps are visualized over one generated image.
}
\label{fig:small}
\end{figure*}

\vspace{-0.3cm}

\section{Introduction}

One of the key realizations of modern machine learning is that models trained on one task end up being useful for many other, often unrelated, tasks.  This is evidenced by the success of backbone pretrained networks and self-supervised training regimes.
In computer vision, the prevailing theory is that neural network models trained for various vision tasks tend to share the same concepts and structures because they are inherently present in the visual world.
 However, the precise nature of these shared elements and the technical mechanisms that enable their transfer remain unclear.  

\blfootnote{Project page, code and models: \href{https://yossigandelsman.github.io/rosetta_neurons/index.html}{\url{https://yossigandelsman.github.io/rosetta_neurons}}}
In this paper, we seek to identify and match units that express similar concepts across different models. We call them {\em Rosetta \footnote{The Rosetta Stone is an ancient Egyptian artifact, a large stone inscribed with the same text in three different languages. It was the key to deciphering Egyptian hieroglyphic script. The original stone is on public display at the British Museum in London.}
Neurons} (see fig.~\ref{fig:teaser}).
How do we find them, considering it is likely that each model would express them differently? 
Additionally, neural networks are usually over-parameterized, which suggests that multiple neurons can express the same concept (synonyms). The layer and channel that express the concept would also differ between models. Finally, the value of the activation is calibrated differently in each. To address these challenges, we carefully choose the matching method we use. We found that post ReLU/GeLU values tend to produce distinct activation maps, thus these are the values we match. We compare units from different layers between the models while carefully normalizing the activation maps to overcome these differences. To address synonym neurons, we also apply our matching method on a model with itself and cluster units together according to the matches.

We search for Rosetta Neurons across eight different models: Class Supervised-ResNet50~\cite{He2015DeepRL}, DINO-ResNet50, DINO-ViT~\cite{dino}, MAE~\cite{mae}, CLIP-ResNet50~\cite{clip}, BigGAN~\cite{biggan}, StyleGAN-2~\cite{stylegan2}, StyleGAN-XL~\cite{styleganxl}. We apply the models to the same dataset and correlate different units of different models. We mine the Rosetta neurons by clustering the highest correlations. This results in the emergence of model-free global representations, dictated by the data.

Fig.~\ref{fig:all_neurons} shows an example image and all the activation maps from the discovered Rosetta Neurons. The activation maps include semantic concepts such as the person's head, hand, shirt, and fish as well as non-semantic concepts like contour, shading, and skin tone. In contrast to the celebrated work of Bau \etal on Network Dissection~\cite{bau2019gandissect,netdissect2017}, our method does not rely on human annotations or semantic segmentation maps. Therefore, we allow for the emergence of non-semantic concepts. 

The Rosetta Neurons allow us to translate from one model's ``language'' to another. One particularly useful type of model-to-model translation is from discriminative models to generative models as it allows us to easily visualize the Rosetta Neurons. By applying simple transformations to the activation maps of the desired Rosetta Neurons and optimizing the generator's latent code, we demonstrate realistic edits. Additionally, we demonstrate how GAN inversion from real image to latent code improves when the optimization is guided by the Rosetta Neurons. This can be further used for out-of-distribution inversion, which performs image-to-image translation using a regular latent-to-image GAN. All of these edits usually require specialized training (e.g. ~\cite{epstein2022blobgan,pix2pix,cyclegan}), but we leverage the Rosetta Neurons to perform them with a fixed pre-trained model.

The contributions of our paper  are as follows:
\begin{itemize}
    \itemsep0em 
    \item We show the existence of Rosetta Neurons that share the same concepts across different models and training regimes. 
    \item We develop a method for matching, normalizing, and clustering activations across models. We use this method to curate a dictionary of visual concepts. 
    \item The Rosetta Neurons enables model-to-model translation that bridges the gap between representations in generative and discriminative models.
    \item We visualize the  Rosetta Neurons and  exploit them as handles to demonstrate manipulations to generated images that otherwise require specialized training.
\end{itemize}

\begin{figure*}
\vspace{-0.5cm}
  \centering \includegraphics[width=0.78\textwidth]{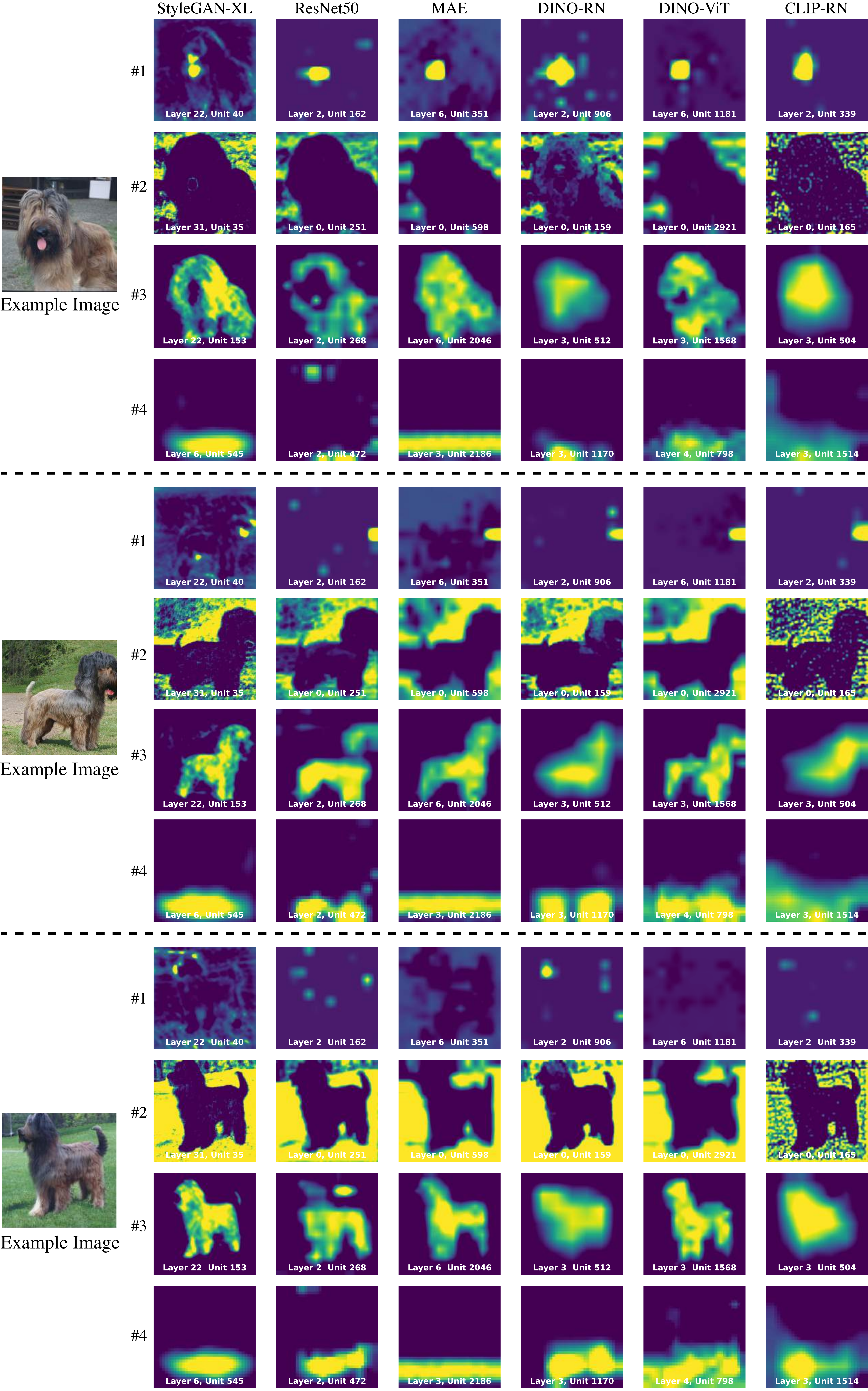}
\caption{\label{fig:huge}
\textbf{Rosetta Neuron Dictionary.} A sample from the dictionary curated for the ImageNet class ``Briard''. The full dictionary can be found in the supplementary material. The figure presents 4 emergent concepts demonstrated in 3 example images. For each model, we present the normalized activation maps of the Rosetta Neuron matching the shared concept. }
\vspace{-0.2cm}
\end{figure*}

\section{Related Work}

\textbf{Visualizing deep representations.}
The field of interpreting deep models has been steadily growing, and includes optimizing an image to maximize the activations of particular neurons~ \cite{Zeiler2013VisualizingAU, springenberg2014striving, olah2020zoom}, gradient weighted activation maps  \cite{SimonyanVZ13,Petsiuk2018rise,thereandback, Selvaraju2016GradCAMVE}, nearest neighbors of deep feature representations \cite{Krizhevsky2012ImageNetCW}, etc. 
The seminal work of Bau \etal \cite{netdissect2017, bau2019gandissect} took a different approach by identifying units that have activation maps highly correlated with semantic segments in corresponding images, thereby reducing the search space of meaningful units. However, this method necessitates annotations provided by a pre-trained segmentation network or a human annotator and is confined to discovering explainable units from a predefined set of classes and in a single model. 
Whereas all previous works focused on analyzing a single, specific neural network model, the focus of our work is in capturing commonalities across many different networks. Furthermore, unlike \cite{bau2019gandissect, netdissect2017}, our method does not require semantic annotation.

\begin{figure*}
\vspace*{-0.5cm}
  \centering
  \hspace*{-0.2cm} \includegraphics[width=0.8\textwidth]{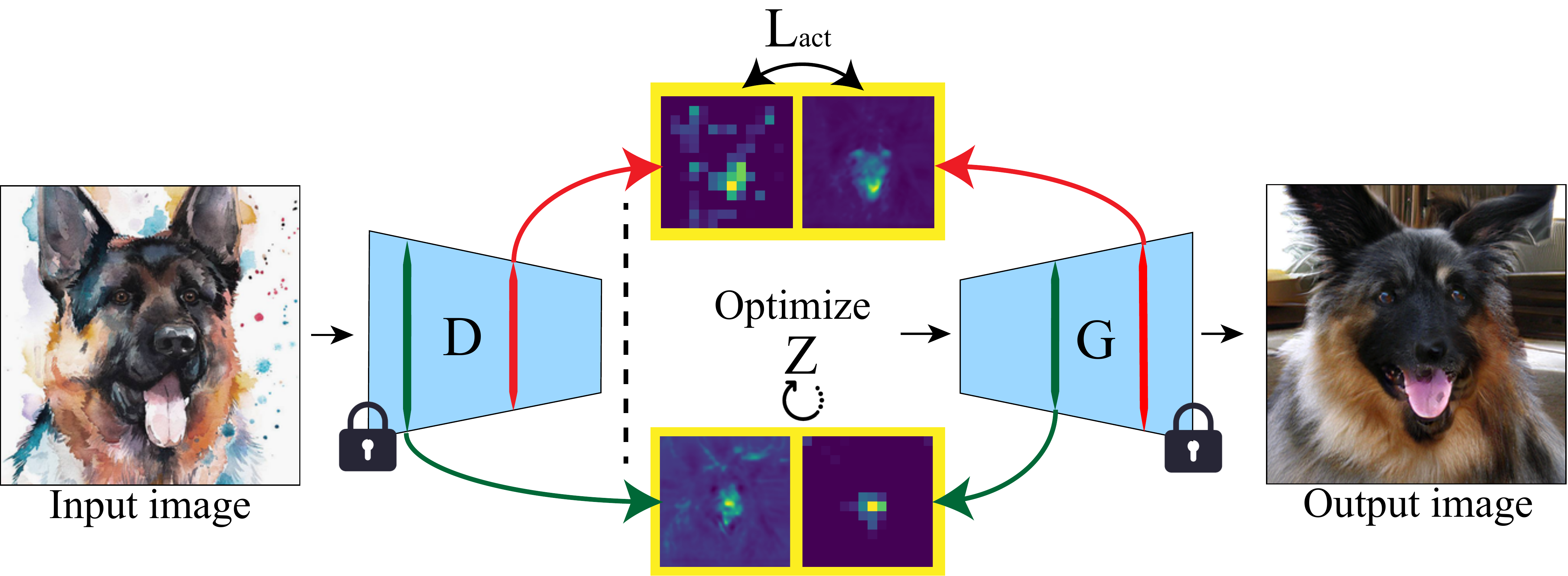}
\caption{\label{fig:inversion_arc}
\textbf{Rosetta Neurons guided image inversion.} An input image is passed through a discriminative model $D$ (i.e.: DINO) to obtain the Rosetta Neurons' activation maps. Then, the latent code $Z$ of the generator is optimized to match those activation maps, according to the extracted pairs. 
}
\vspace{-0.2cm}
\end{figure*}

\textbf{Explaining discriminative models with generative models.} GANAlyze~\cite{GANalyze} optimized the latent code of a pre-trained GAN to find directions that affect a classifier decision. Semantic Pyramid~\cite{semanticpyramid} explored the subspaces of generated images to which the activations of a classifier are invariant. Lang~\etal~\cite{Lang2021ExplainingIS} trained a GAN to explain attributes that underlie classifier decisions. In all of these cases, the point where the generative and discriminative models communicate is in the one ``language'' they both speak - pixels; which is the output of the former and an input of the latter. Our method for bridging this gap takes a more straightforward approach: we directly match neurons from pre-trained networks and identify correspondences between their internal activations. Moreover, as opposed to \cite{Lang2021ExplainingIS} and \cite{semanticpyramid}, our method does not require GAN training and can be applied to any off-the-shelf GAN and discriminative model.

\textbf{Analyzing representation similarities in neural networks}. Our work is inspired by the neuroscience literature on representational similarity analysis \cite{10.3389/neuro.06.004.2008,edelman_1998} that aims to extract correspondences between different brain areas \cite{Haxby}, species~\cite{Kriegeskorte2008MatchingCO}, individual subjects~\cite{Connolly2012TheRO}, and between neural networks and brain neural activities~\cite{yamins}. On the computational side, Kornblith~\etal~\cite{Kornblith2019SimilarityON} aimed to quantify the similarities between different layers of discriminative convolutional neural networks, focusing on identifying and preserving invariances. Esser, Rombach, and Ommer~\cite{Esser2020ADI, Rombach2020NetworktoNetworkTW} trained an invertible network to translate non-local concepts, expressed by a latent variable, across models. In contrast, our findings reveal that individual neurons hold shared concepts across a range of models and training regimes without the need to train a specialized network for translation. This leads to another important difference: the concepts we discover are local and have different responses for different spatial locations in an image. We can visualize these responses and gain insights into how these concepts are represented in the network.


\section{Method}
Our goal is to find Rosetta Neurons across a variety of models. We define Rosetta Neurons as two (or more) neurons in different models whose activations (outputs) are positively correlated over a set of many inputs.
Below we explain how to find Rosetta Neurons across a variety of models and describe how to merge similar Rosetta Neurons into clusters that represent the same concepts.

\subsection{Mining common units in two models}
\textbf{Preliminaries.}
Given two models $F^{(1)}, F^{(2)}$, we run $n$ inputs through both models. For discriminative models, this means a set of images ${\{I_i\}^n_{i=1}}$. If one of the models is generative, we first sample $n$ random input noises ${\{Z_i\}^n_{i=1}}$ and generate images $I_i=F^{(1)}(z_i)$ that will be the set of inputs to the discriminative model $F^{(2)}$. We denote the set of extracted activation maps of $F$ by $F^{act}$. The size $|F^{act}|$ is the total number of channels in all the layers. The $j$-th intermediate activation map of $F$ when applied to the $i$-th input is then $F^j_i$. That is $F^j_i=F^j(I_i)$ for a discriminative model and $F^j_i=F^j(z_i)$ for a generative one.






\textbf{Comparing activation maps.} 
To compare units $F^{(1)j}$ and $F^{(2)k}$, namely, the $j$-th unit from the first model with the $k$-th unit from the second one, we first bilinearly interpolate the feature maps to have the same spatial dimensions according to the maximum of the two map sizes.
Our approach to perform matching is based on correlation, similar to \cite{10.3389/neuro.06.004.2008}, but taken across both data instances and spatial dimensions. 
We then take the mean and variance across the $n$ images and across the spatial dimensions of the images, where $x$ combines both spatial dimensions of the images.
\small
\begin{equation}\label{eq1}\begin{split}
\overline{F^j} &= \frac{1}{nm^2}\sum\limits_{i,x} F^j_{i,x}\\
var(F^j) &= \frac{1}{nm^2-1}\sum\limits_{i,x} \left(F^j_{i,x} - \overline{F^j}\right)^2
\end{split}\end{equation}
\normalsize  
Next, the measure of distance between two units is calculated by Pearson correlation:
\small
\begin{equation}
d(F^{(1)j},F^{(2)k})=\frac{\sum\limits_{i,x} 
\left(F^{(1)j}_{i,x} - \overline{F^{(1)j}} \right) 
\left(F^{(2)k}_{i,x} - \overline{F^{(2)k}} \right) }
{\sqrt{
var(F^{(1)j}) \cdot var(F^{(2)k})
}}
\end{equation}

\normalsize
\
\normalsize
In our experiments, this matching is computed between a generative model $G$ and a discriminative model $D$. The images used for $D$ are generated by $G$ applied to $n$ sampled noises.

\textbf{Filtering ``best buddies" pairs.} To detect reliable matches between activation maps, we keep the pairs that are mutual nearest neighbors (named ``best-buddies'' pairs by \cite{bestbuddies}) according to our distance metric and filter out any other pair. Formally, our set of ``best buddies'' pairs is: 
\vspace{-0.2cm}
\small
\begin{equation}
\begin{split}
    BB(&F^{(1)}, F^{(2)}; K) =\{(j, k)|\\
             & F^{(1)k} \in KNN(F^{(2)j},F^{(1)act}; K) \\
    \land \;\; & F^{(2)j} \in KNN(F^{(1)k},F^{(2)act}; K) \} \\
\end{split}
\end{equation}
\vspace{-0.25cm}

\noindent Where $ KNN({F^{(a)j},F^{(b)act}}) $ is the {set of the} {K}-nearest neighbors {of the unit 
 ${j}$ from model ${F^{(a)}}$ among all the units in model ${F^{(b)}}$}:
 \vspace{-0.25cm}
\begin{equation*}
\begin{split}
KNN(F^{(a)j}, F^{(b)act}; K) = &\underset{{q_1...q_K} \subseteq F^{(b)act}}{\mathrm{argmin}} \sum_{k=1}^{K} d(F^{(a)j}, q_k)
\end{split}
\vspace{-0.2cm}
\end{equation*} As shown in~\cite{bestbuddies}, the probability of being mutual nearest neighbors is maximized when the neighbors are drawn from the same distribution. Thus, keeping the ``best buddies'' discards noisy matches.


\subsection{Extracting common units in $m$ models}

\textbf{Merging units between different models.} To find similar activation maps across  many different discriminative models $D_i, i\in [m]$, we merge the ``best buddies'' pairs calculated between $D_i$ and a generator $G$ for all the $i$'s. Formally, our Rosetta units are:
\small
\begin{equation}
\begin{split}    
    R(G, D_1...D_m) = \{(j,k_1,...,k_m)| \forall{i}:(j,k_i) \in BB(G,D_i)\}
\end{split}
\end{equation}
\normalsize
This set of tuples includes the ``translations'' between similar neurons across all the models. Note that when $m=1$, $R(G,D_1)=BB(G,D_1)$.

\textbf{Clustering similar units into concepts.}
Empirically, the set of Rosetta units includes a few units that have similar activation maps for the $n$ images. For instance, multiple units may be responsible for edges or concepts such as ``face.'' We cluster them according to the self ``best-buddies" of the generative model, defined by $BB(G, G;K)$.
We set two Rosetta Neurons in $R$ to belong to the same cluster if their corresponding units in $G$ are in $BB(G, G;K)$.

\textbf{Curating a dictionary.} After extracting matching units for a dataset across a model zoo, we enumerate the sets of matching Rosetta Neurons in the clustered $R$. Fig.~\ref{fig:huge} is a sample from such a dictionary. Fig.~\ref{fig:small} shows a list of all the concepts for a single image.
Since the concepts emerge and are not related to human annotated labels, we simply enumerate them and present each concept on several example images to visually identify it. Using 1600 instances generated by the GAN, Distances are taken between all possible bipartite pairs of units, the $K=5$ nearest neighbors are extracted, from which Best-Buddies are filtered. Typically for the datasets and models we experimented with, around 50 concepts emerge. The exact list of models used in our experiments and the datasets they were trained on can be found in Table.~\ref{tab:models}. See supplementary material for the dictionaries.

\section{Visualizing the Rosetta Neurons}
As we involve a generative model in the Rosetta Neurons mining procedure, we can utilize it for visualizing the discovered neurons as well. In this section, we present how to visualize the neurons via a lightweight matches-guided inversion technique. We then present how direct edits of the activation maps of the neurons can translate into a variety of generative edits in the image space, without any generator modification or re-training.

\begin{figure}
\vspace*{-0.4cm}
  \centering
  \hspace*{-0.2cm} \includegraphics[width=\columnwidth]{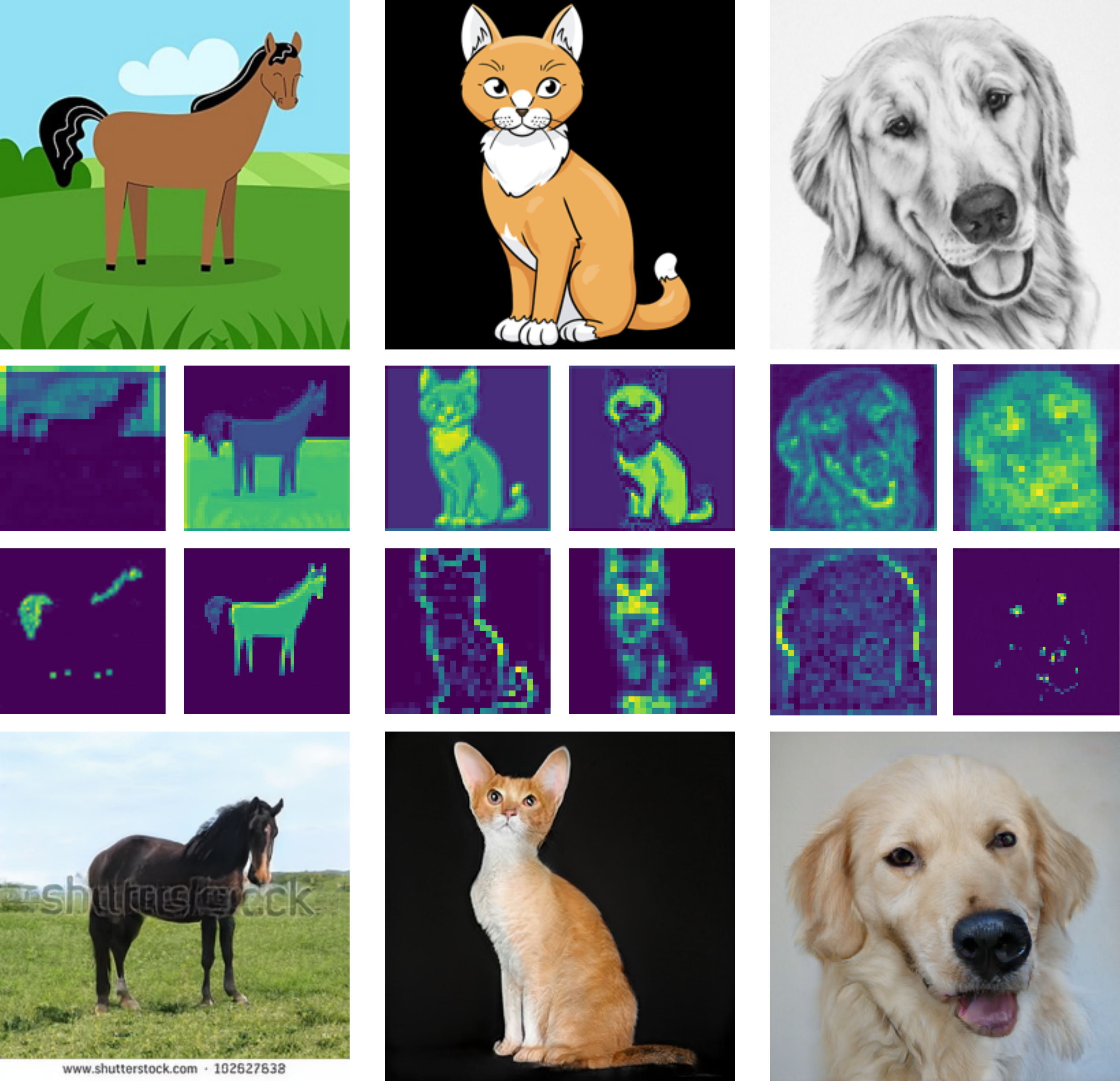}
\caption{\label{fig:ood}
\textbf{Out-of-distribution inversions}. By incorporating the Rosetta Neurons in the image inversion process, we can invert sketches and cartoons (first row), and generate similar in-distribution images (last row). A subset of the Rosetta Neurons from the input images that were matched during the inversion process is shown in the middle rows.
}
\end{figure}

\subsection{Rosetta Neurons-Guided Inversion}
To visualize the extracted Rosetta Neurons, we take inspiration from \cite{semanticpyramid}, and use the generative model $G$ to produce images for which the generator activation maps of the Rosetta Neurons best match to the paired activation maps extracted from $D(I_v)$, as shown in figure \ref{fig:inversion_arc}. As opposed to \cite{semanticpyramid}, we do not train the generative model to be conditioned on the activation maps. Instead, we invert images through the fixed generator into some latent code $z$, while maximizing the similarity between the activation maps of the paired Rosetta Neurons. Our objective is: 
\begin{equation}
    \arg \min_{z} ( - L_{act}(z,I_v) + \alpha L_{reg} (z)) 
\label{direct_opt1}
\end{equation}  
Where $\alpha$ is a loss coefficient, $L_{reg}$ is a regularization term ($L_2$ or $L_1$), and $L_{act}(z,I_v)$ is the mean of normalized similarities between the paired activations:
\small
\begin{equation}\begin{split}
&L_{act}(z,I_v) =\\
&\frac{1}{|BB(G,D)|}\mathlarger{\mathlarger{\sum}}_{\substack{(j,k) \in \\ BB(G,D)}}
\frac{\sum\limits_{x} 
\left(G^j_x - \overline{G^j} \right) 
\left(D^k_x - \overline{D^k} \right) }
{\sqrt{
var(G^j) \cdot var(D^k)
}}
\label{inv_loss}
\end{split}\end{equation}
\normalsize
Where $G^j$ is the $j$-th activation map of $G(z)$ and $D^k$ is the $k$-th activation map of $D(I_v)$.
For obtaining this loss, we use the mean and variance precomputed by Eq.~\ref{eq1} over the entire dataset during the earlier mining phase. However, we calculate the correlation over the spatial dimensions of a single data instance. 

\begin{figure}
\vspace*{-0.4cm}
  \centering
  \hspace*{-0.2cm} \includegraphics[width=\columnwidth]{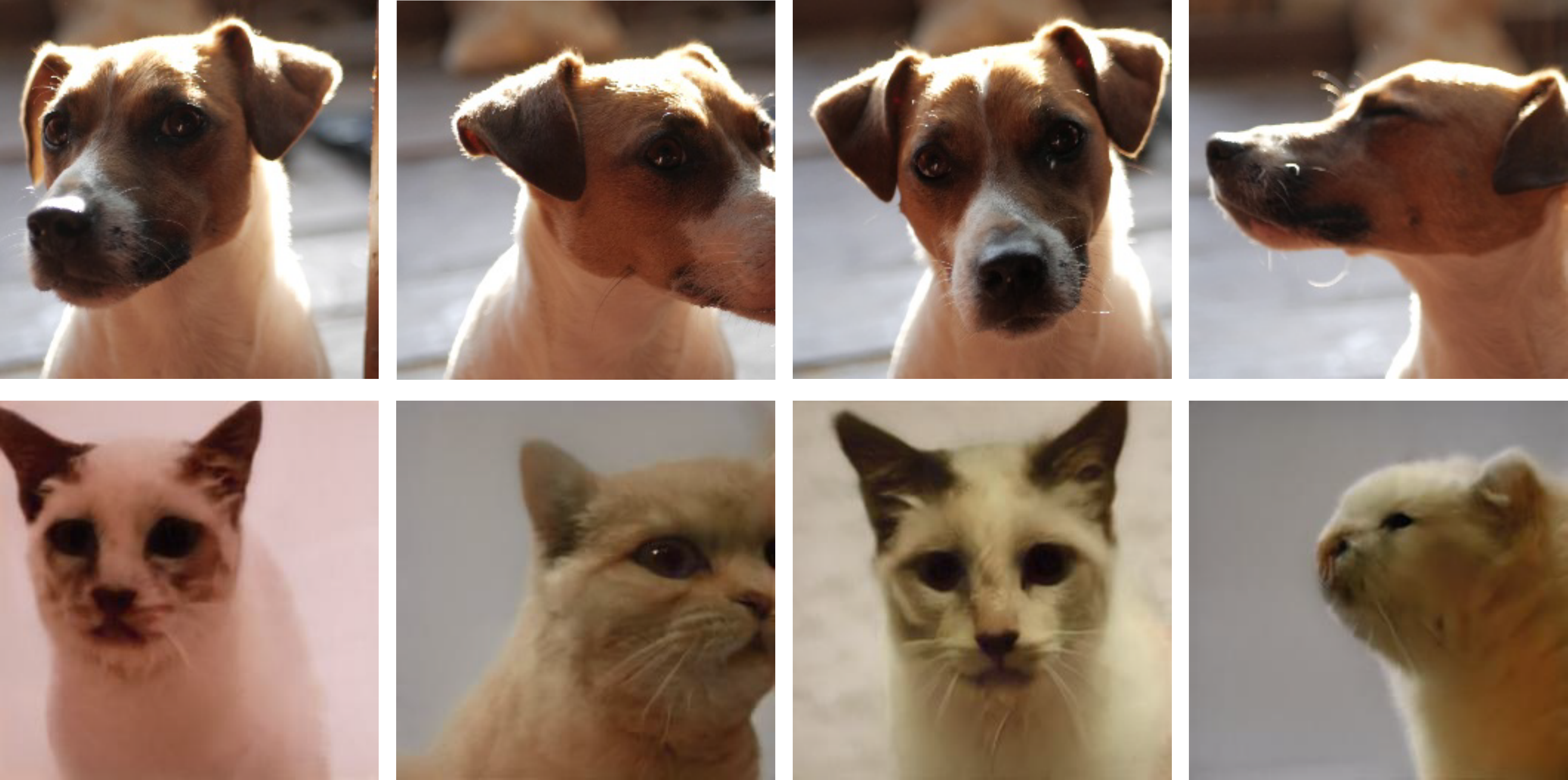}
\caption{\label{fig:pose}
\textbf{Cross-class image-to-image translation.} Rosetta Neurons guided inversion of input images (top row) into a StyleGAN2 trained on LSUN cats \cite{yu15lsun}, allows us to preserve the pose of the animal while changing it from dog to cat (bottom row). See supplementary material for more examples. 
}
\end{figure}

The Rosetta neurons guided inversion has two typical modes. The first mode is when both the initial activation map and the target one have some intensity somewhere in the map (e.g. two activation maps that are corresponding to ``nose'' are activated in different spacial locations). In this case, the visual effect is an alignment between the two activation maps. As many of the Rosetta neurons capture object parts, it results in image-to-image alignment (e.g., fig.~\ref{fig:pose}). The second mode is when either the target or the initial activation map is not activated. In this case, a concept will appear or disappear (e.g., fig.~\ref{fig:removals}).

\textbf{Visualizing a single Rosetta Neuron.} We can visualize a single Rosetta Neuron by modifying the loss in our inversion process (eq. \ref{inv_loss}). Rather than calculating the sum over the entire set of Rosetta Neurons, we do it for a single pair that corresponds to the specific Rosetta neuron. When this optimization procedure is applied a few times on the same input neuron pair starting from a few different randomly initialized latent codes, we get a diverse set of images that are matching to the same activation map of the wanted Rosetta Neuron. This allows a user to disentangle and detect what is the concept that is specifically represented by the given neuron. Figure \ref{fig:teaser} present two optimized images for each of the presented Rosetta Neurons. This visualization allows the viewer to see that Concept \#1 corresponds to the concept ``red color,'' rather than to the concept ``hat.''


\textbf{Inverting out-of-distribution images.} The inversion process presented above does not use the generated image in the optimization, as opposed to common inversion techniques that calculate the pixel loss or perceptual loss between the generated image the input image. Our optimization process does not compare the image pixel values, and as many of the Rosetta Neurons capture high-level semantic concepts and coarse structure of the image, this allows us to invert images outside of the training distribution of the generative model. Figure \ref{fig:pose} presents a cross-class image-to-image translation that is achieved by Rosetta Neurons guided inversion. As shown, the pose of the input images of dogs is transferred to the poses of the optimized cat images, as the Rosetta Neurons include concepts such as ``nose,'' ``ears,'' and ``contour'' (please refer to Figure \ref{fig:teaser} for a subset of the Rosetta Neurons for this set of models).  

Figure \ref{fig:ood} presents the inversion results for sketches and cartoons, and a subset of the Rosetta Neurons that were used for optimization. As shown, the matches-guided inversion allows us to ``translate'' between the two domains via the shared Rosetta Neurons and preserve the scene layout and object pose. Our lightweight method does not require dedicated models or model training, as opposed to \cite{cyclegan,pix2pix}.


\textbf{Inverting in-distribution images.} We found that adding the loss term in eq. \ref{direct_opt1} to the simple reconstruction loss objective improves the inversion quality. Specifically, we optimize:
\begin{equation}
    \arg \min_{z} (L_{rec} (G(z), I_v) + \alpha L_{reg} (z) - \beta L_{act}(z, I_v)) 
\label{direct_opt}
\end{equation}
Where $L_{rec}$ is the reconstruction loss between the generated image and the input image,  and $\beta$ is a loss coefficient. The reconstruction loss can be pixel loss, such as $L_1$ or $L_2$ between the two images, or a perceptual loss.

 We compare the inversion quality with and without the Rosetta Neurons guidance and present the PSNR, SSIM, and LPIPS \cite{zhang2018perceptual} for StyleGAN-XL inversion. We use solely a perceptual loss as a baseline, similarly to \cite{styleganxl}. We add our loss term to the optimization, where the Rosetta Neurons are calculated from 3 sets of matches with StyleGAN-XL: matching to DINO-RN, matching to CLIP-RN, and matching across all the discriminative models in Table \ref{tab:models}.  We use the same hyperparameters as in \cite{styleganxl}, and set $\alpha = 0.1$ and $\beta = 1$.
 
\begin{figure}
\vspace*{-0.4cm}
  \centering
  \hspace*{-0.2cm} \includegraphics[width=0.9\columnwidth]{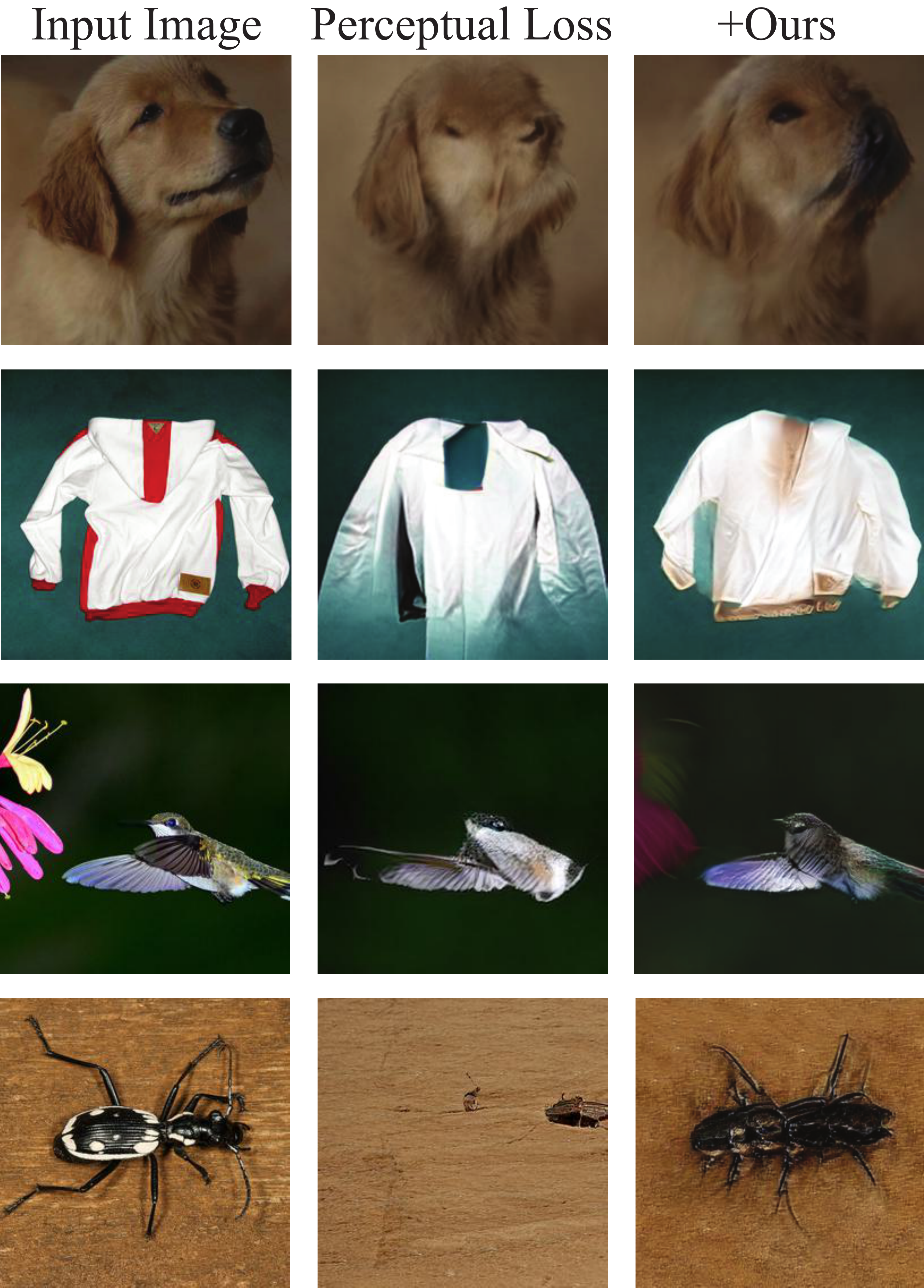}
\caption{\label{fig:inversion}
\textbf{Image inversions for StyleGAN-XL.} We compare inversions by optimizing perceptual loss only (second column), to additional Rosetta Neurons guidance loss, with matches calculated across all the models presented in Figure \ref{fig:huge} (third column). See supplementary material for more examples. \vspace{-0.4cm}
}
\end{figure}

Table \ref{tab-inv} presents the quantitative inversion results for 5000 randomly sampled images from the ImageNet validation set (10\% of the validation set, 5 images per class), as done in \cite{styleganxl}. Figure \ref{fig:inversion} presents the inversion results for the baseline and for the additional Rosetta Neurons guidance using the matches between all the models. As shown qualitatively and quantitatively, the inversion quality improves when the Rosetta Neurons guiding is added. We hypothesize this is due to the optimization objective that directly guides the early layers of the generator and adds layout constraints. These soft constraints reduce the optimization search space and avoid convergence to local minima with low similarity to the input image.

\begin{table}
\vspace{-0.3cm}
\centering
\scalebox{0.9}{
\begin{tabular}{l|c|c|c}

& \textbf{PSNR 	$\uparrow$} & \textbf{SSIM 	$\uparrow$} & \textbf{LPIPS 	$\downarrow$}\\
\hline
Perceptual loss & 13.99 & 0.340 & 0.48 \\ 
+DINO matches & 15.06 & 0.360 & 0.45 \\ 
+CLIP matches & 15.20 & 0.362 & \textbf{0.44} \\ 
+All matches & \textbf{15.42} & \textbf{0.365} & 0.46 \\ 
\hline
\end{tabular}}
\caption{\small
\textbf{Inversion quality on ImageNet.} We compare the inversion quality for StyleGAN-XL when Rosetta Neurons guidance is added, for 3 sets of matches -  StyleGAN-XL \& DINO-RN, StyleGAN-XL \& CLIP-RN and all the models from figure \ref{fig:huge}. 
}
\label{tab-inv}
\end{table}

\begin{table}
\centering
\scalebox{0.8}{
\begin{tabular}{c|c|c}
Model & Training dataset & Resolution\\
\hline
StyleGAN-XL & ImageNet & 256 \\
StyleGAN2 & LSUN(cat) & 256 \\
StyleGAN2 & LSUN(horse) & 512 \\
BigGAN & ImageNet & 256 \\ 
ResNet50 & ImageNet & 224 \\
DINO-ResNet50 & ImageNet & 224 \\ 
DINO-VIT-base & ImageNet & 224 \\ 
MAE-base & ImageNet & 224 \\
CLIP & WebImageText & 224 \\
\hline
\end{tabular}
}
\vspace{-0.3cm}
\caption{
\textbf{Models used in the paper.}
\vspace{-0.5cm}
}
\label{tab:models}
\end{table}

\normalsize


\subsection{Rosetta Neurons Guided Editing}
The set of Rosetta Neurons allows us to apply controlled edits on a generated image $I_{src} = G(z)$ and thus to provide a counterfactual explanation to the neurons. Specifically, we modify the activation maps corresponding to the Rosetta Neurons, extracted from $G(z)$, and re-optimize the latent code to match the edited activation maps according to the same optimization objective presented in eq. \ref{direct_opt1}. As opposed to previous methods like \cite{epstein2022blobgan}, which trained a specifically designed generator to allow disentangled manipulation of objects at test-time, we use a fixed generator and only optimize the latent representation. Next, we describe the different manipulation that can be done on the activation maps, before re-optimizing the latent code:

\textbf{Zoom-in.} We double the size of each activation map that corresponds to a Rosetta Neurons with bilinear interpolation and crop the central crop to return to the original activation map size. We start our re-optimization from the same latent code that generated the original image.  

\textbf{Shift.} To shift the image, we shift the activation maps directly and pad them with zeros. The shift stride is relative to the activation map size (e.g. we shift a $4~\times~4$ activation map by 1, while shifting $8 \times 8$ activation maps by 2).

\textbf{Copy \& paste.} We shift the activation maps twice into two directions (e.g. left and right), creating two sets of activation maps - left map, and right map. We merge them by copying and pasting the left half of the left activation map and the right half of the right activation map. We found that starting from random $z$ rather than $z$ that generated the original image obtains better results.

\begin{figure}
  \centering
  \includegraphics[width=\columnwidth]{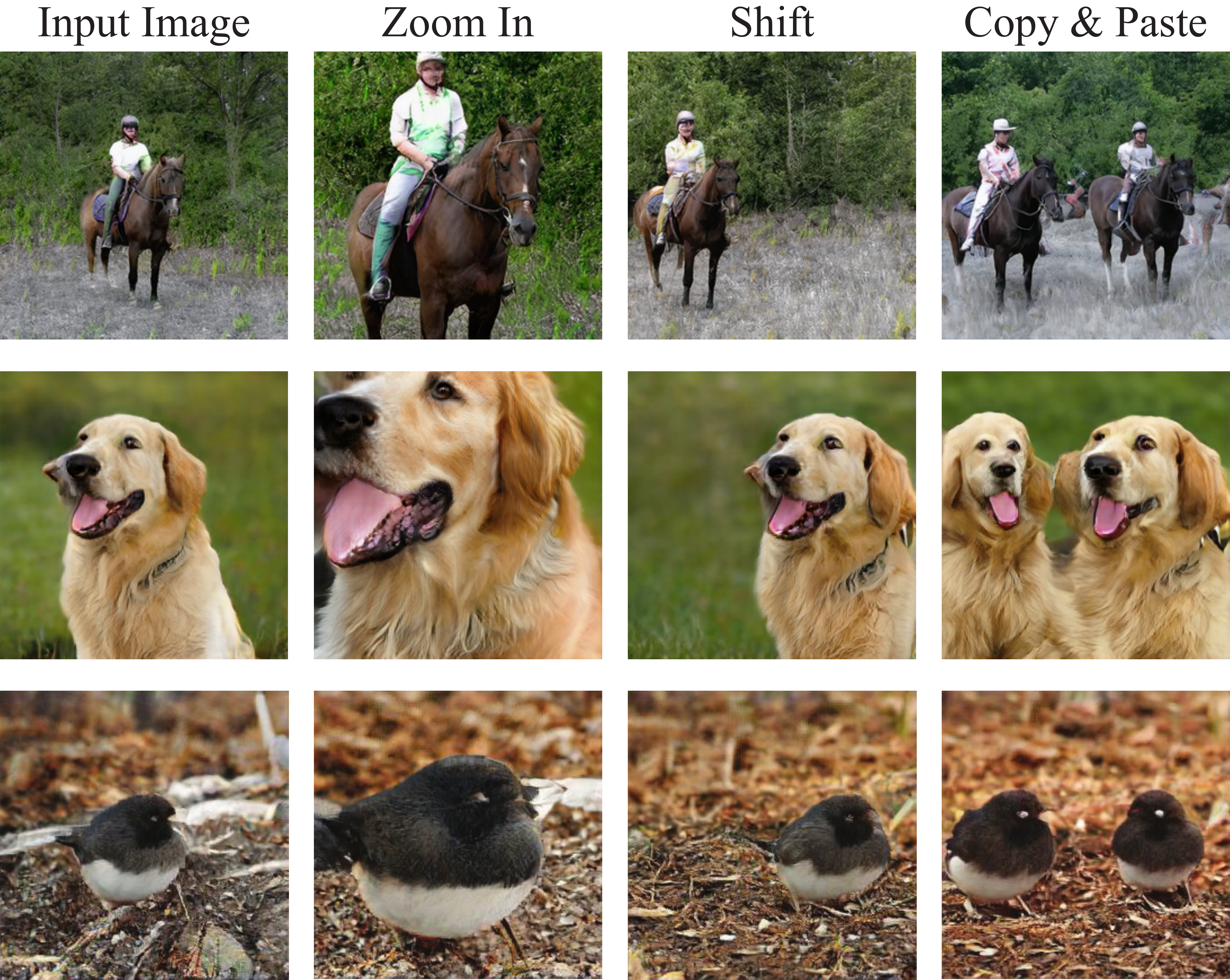}
  \vspace{-0.5cm}
\caption{\label{fig:editing}
\textbf{Rosetta Neurons guided editing}. Direct manipulations on the activation maps corresponding to the Rosetta neurons are translated to manipulations in the image space. We use two models (top row - StyleGAN2, bottom two rows - BigGAN) and utilize the matches between each of them to DINO-RN. 
\vspace{-0.4cm}
}
\end{figure}

Figure \ref{fig:editing} shows the different image edits that are done via latent optimization to match the manipulated Rosetta Neurons. We apply the edits for two different generative models (BigGAN and StyleGAN2) to show the robustness of the method to different architectures.

\begin{figure}
  \centering
  \hspace*{-0.2cm} \includegraphics[width=\columnwidth]{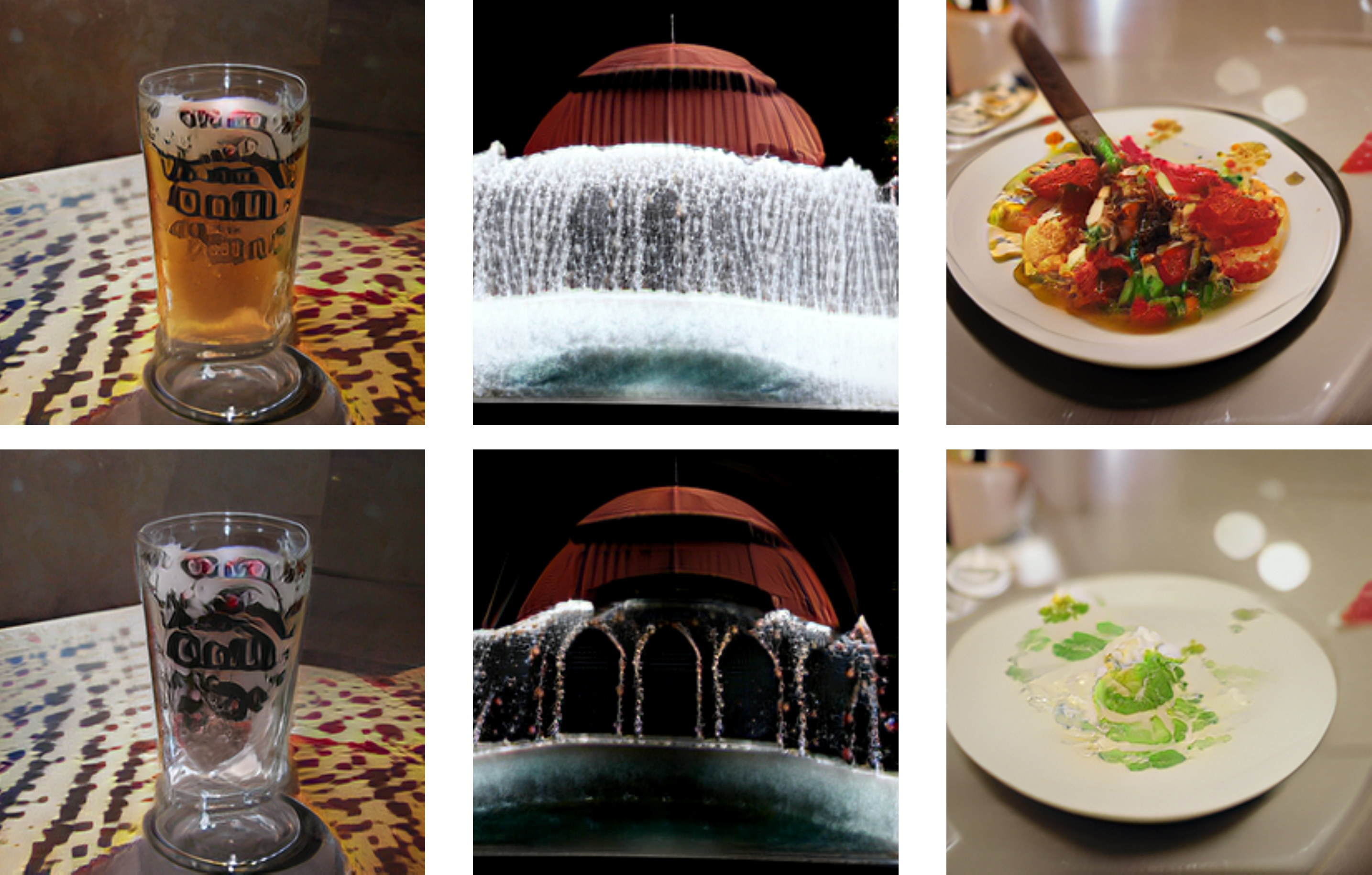}
  \vspace{-0.2cm}
\caption{\label{fig:removals}
\textbf{Single Rosetta Neurons Edits.} 
We optimize the latent input s.t. the value of a desired Rosetta activation reduces. This allows removing elements from the image (e.g. emptying the beer in the glass, reducing the water stream in the fountain, and removing food from a plate). See appendix for more examples.
\vspace{-0.4cm}
}
\end{figure}

\textbf{Fine-grained Rosetta Neurons edit.} Our optimization procedure allows us to manipulate a subset of the Rosetta Neurons, instead of editing all of the neurons together. Specifically, we can manually find among the Rosetta Neurons a few that correspond to elements in the image that we wish to modify. We create ``ground truth" activations by modifying them manually and re-optimizing the latent code to match them. For example - to remove concepts specified by Rosetta Neurons, we set their values to the minimal value in their activation maps. We start our optimization from the latent that corresponds to the input image and optimize until the picked activation maps converge to the manually edited activation maps. Figure \ref{fig:removals} presents examples of removed Rosetta Neurons. Modifying only a few activation maps (1 or 2 in the presented images) that correspond to the objects we aimed to remove, allows us to apply realistic manipulations in the image space. As opposed to \cite{bau2019gandissect}, we do not rewrite the units in the GAN directly and apply optimization instead, as we found that direct edits create artifacts in the generated image for large and diverse GANs.

\textbf{Implementation details.} For the re-optimization step, we train $z$ for 500 steps, with Adam optimizer \cite{adam} and a learning rate of 0.1 for StyleGAN2 and 0.01 for BigGAN. Following \cite{styleganxl}, the learning rate is ramped up from zero linearly during the first 5\% of the iterations and ramped down to zero using a cosine schedule during the last 25\%
of the iterations. We use $K=5$ for calculating the nearest neighbors. The inversion and inversion-based editing take less than 5 minutes per image on one A100 GPU.

\section{Limitations}
Our method can not calculate GAN-GAN matches directly, only through a discriminative model. Unlike discriminative models that can receive the same input image, making two GANs generate the same image is not straightforward. 
Consequently, we only match GANs with discriminative models. 

Secondly, we were unsuccessful when applying our approach to diffusion models, such as \cite{rombach2021highresolution}. We speculate that this is due to the autoregressive nature of diffusion models, where each step is a conditional generative model from image to image. We hypothesize that as a result, the noisy image input is a stronger signal in determining the outcome of each step, rather than a specific unit. Thus, the units in diffusion models have more of an enhancing or editing role, rather than a generating role, which makes it less likely to identify a designated perceptual neuron. 

Lastly, our method relies on correlations, and therefore there is a risk of mining spurious correlations. As shown in Figure \ref{fig:huge}, the dog in the third example does not have its tongue visible, yet both StyleGAN-XL and DINO-RN activated for Concept \#1 in a location where the tongue would typically be found. This may be due to the correlation between the presence of a tongue and the contextual information where it usually occurs.


\section{Conclusion}
We introduced a new method for mining and visualizing common representations that emerge in different visual models. Our results demonstrate the existence of specific units that represent the same concepts in a diverse set of deep neural networks, and how they can be utilized for various generative tasks via a lightweight latent optimization process. We believe that the found common neurons can be used in a variety of additional tasks, including image retrieval tasks and more advanced generative tasks. Additionally, we hope that the extracted representations will shed light on the similarities and dissimilarities between models that are trained for different tasks and with different architectures. We plan to explore this direction in future work.

{\small
\section*{Acknowledgements}
\noindent The authors would like to thank Niv Haim, Bill Peebles, Sasha Sax, Karttikeya Mangalam, and Xinlei Chen for the helpful discussions. YG is funded by the Berkeley Fellowship. AS gratefully acknowledges financial support for this publication by the Fulbright U.S. Postdoctoral Program, which is sponsored by the U.S. Department of State. Its contents are solely the responsibility of the author and do not necessarily represent the official views of the Fulbright Program or the Government of the United States.
Additional funding came from DARPA MCS and ONR MURI.
\bibliographystyle{ieee_fullname}
\bibliography{egpaper_final}
}
\clearpage

\begin{figure*}[hbt!]
\section{Appendix}
  \centering 
  We provide extended examples of Rosetta dictionaries as well as additional edits and visualizations. We further provide the code for extracting and visualizing Rosetta neurons.

  \vspace{0.5cm}
  \includegraphics[width=0.75\textwidth]{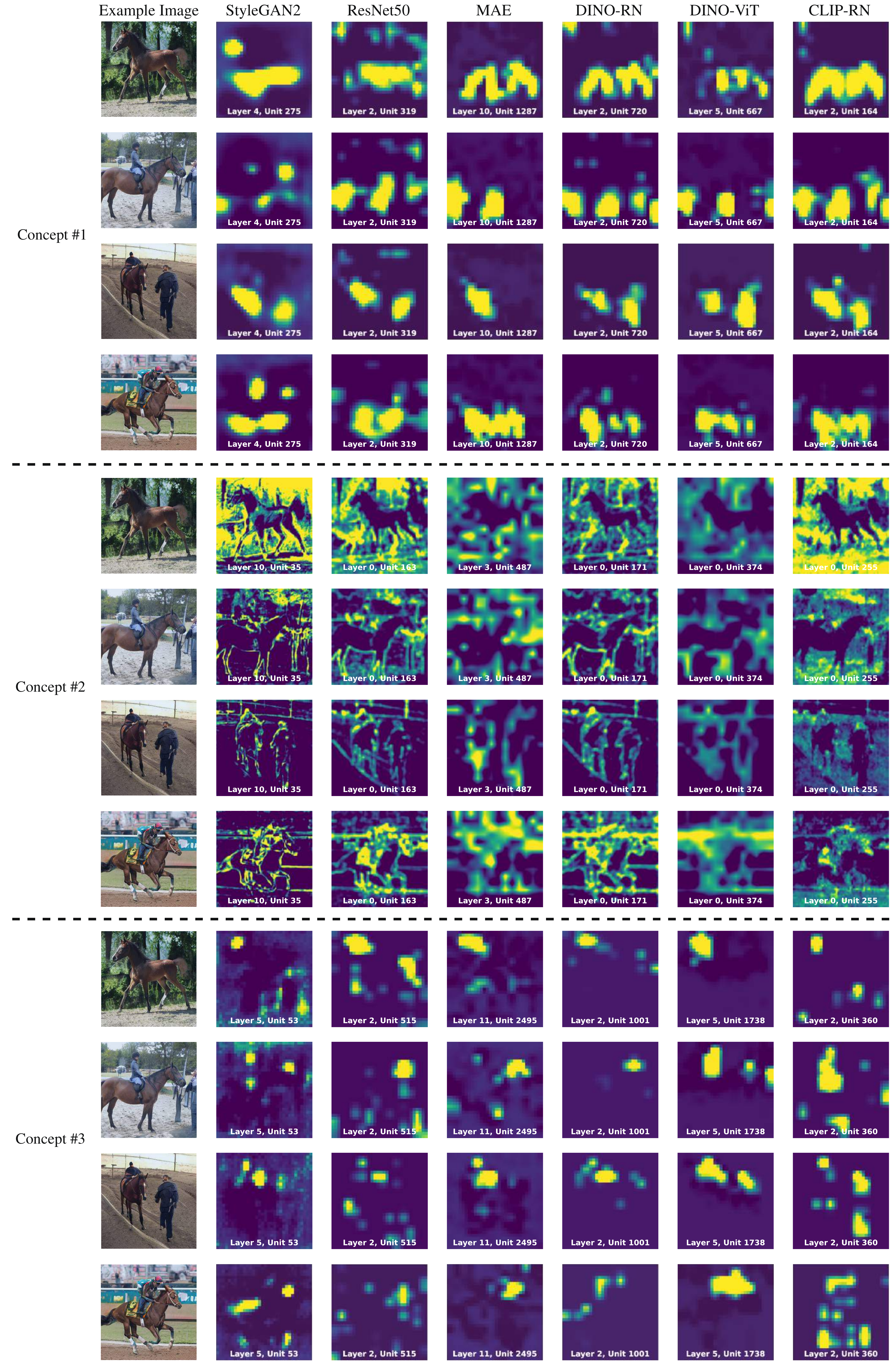}
\caption{\label{fig:sup1}
\textbf{Rosetta Neuron Dictionary for LSUN-horses.} A sample from the dictionary curated for the LSUN-horses dataset. The figure presents 6 emergent concepts demonstrated in 4 example images. }
\end{figure*}
\begin{figure*}[!b]
  \centering \includegraphics[width=0.75\textwidth]{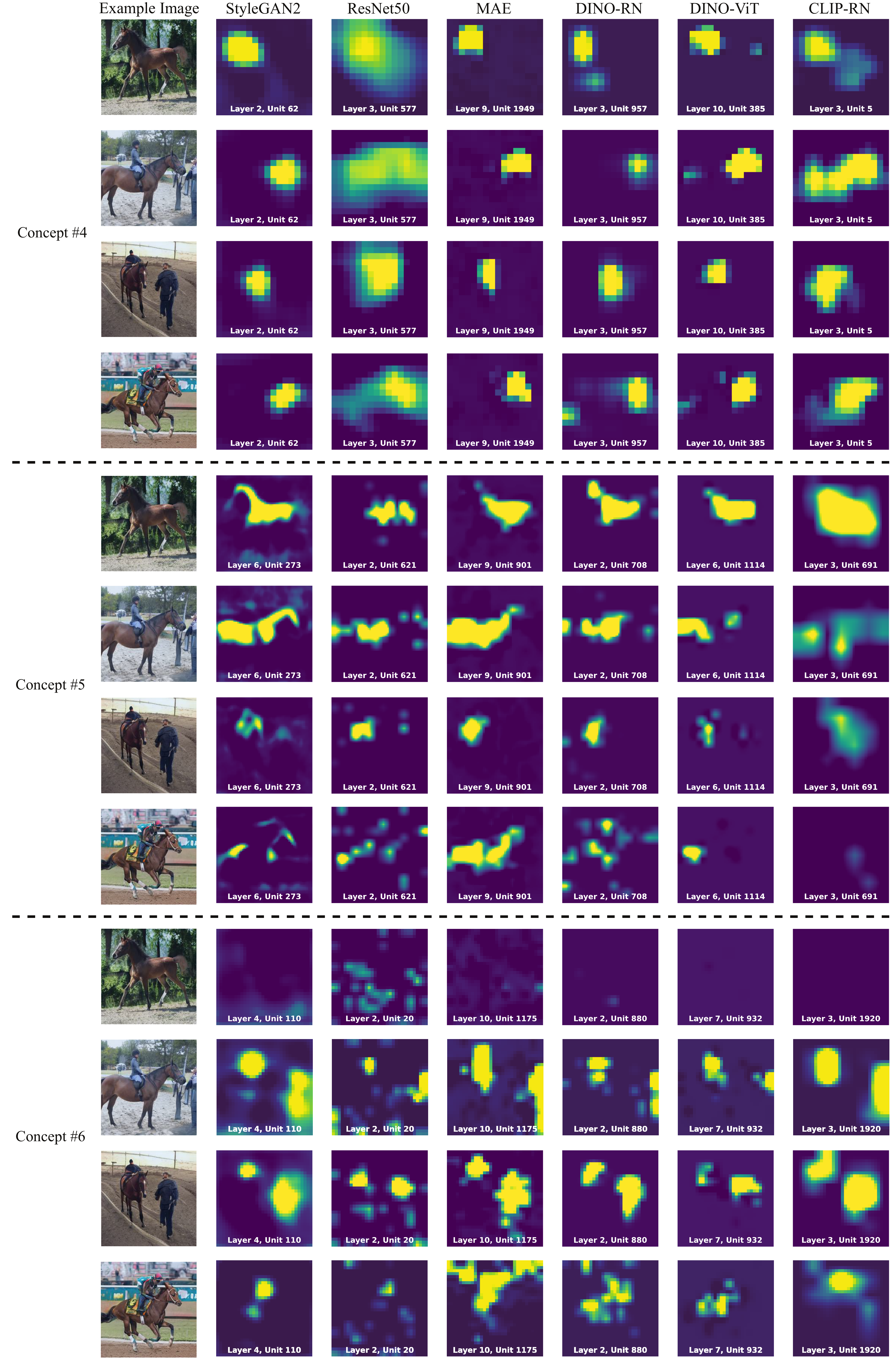}
\caption{\label{fig:sup2}
\textbf{Rosetta Neuron Dictionary for LSUN-horses (cont.)}}
\end{figure*}
\begin{figure*}[!b]
  \centering \includegraphics[width=0.75\textwidth]{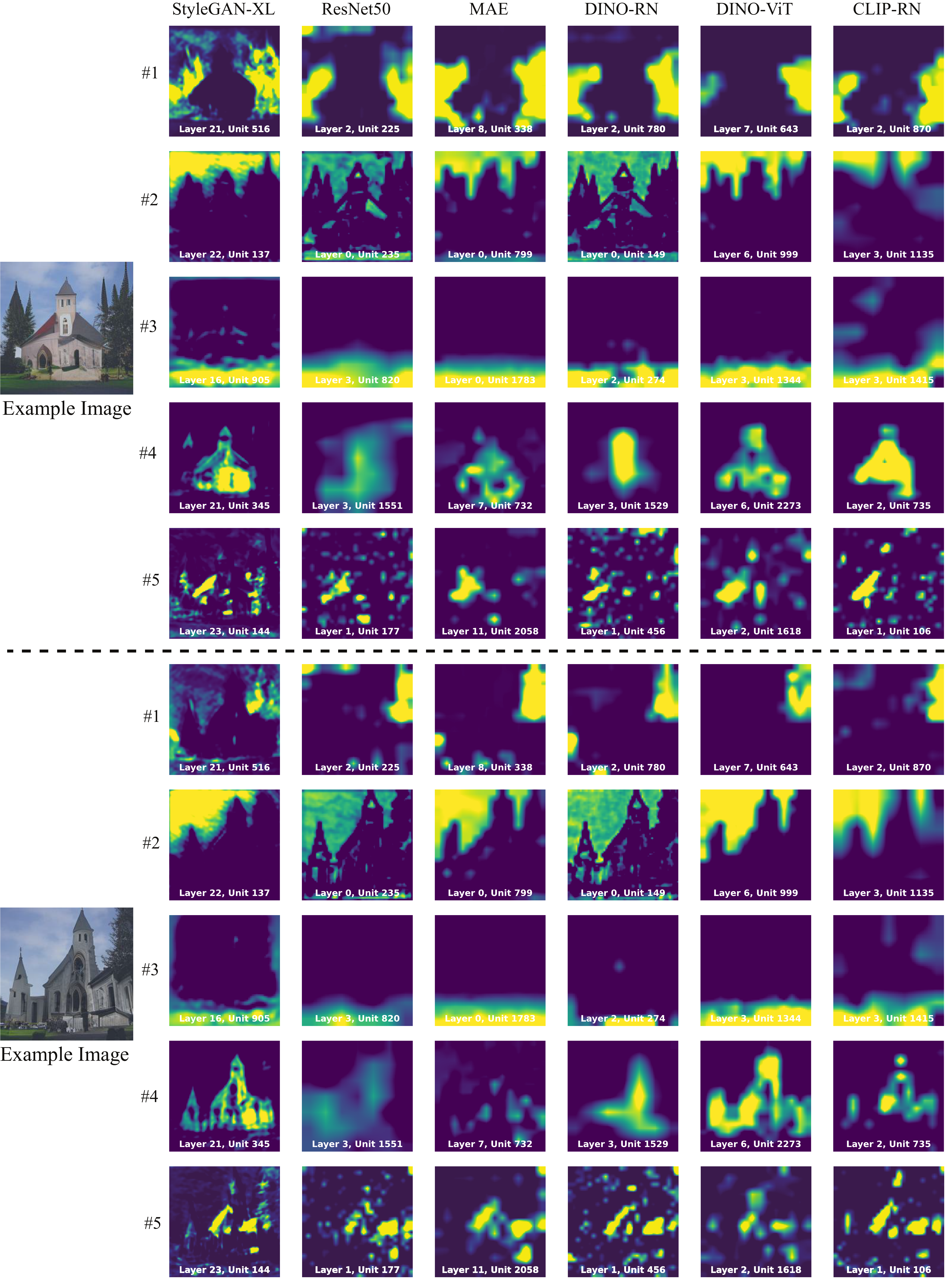}
\caption{\textbf{Rosetta Neuron Dictionary.} A sample from the dictionary curated for the ImageNet class ``Church''. The figure presents 5 emergent concepts demonstrated in 2 example images.}
\end{figure*}

\begin{figure*}
  \centering \includegraphics[width=\textwidth]{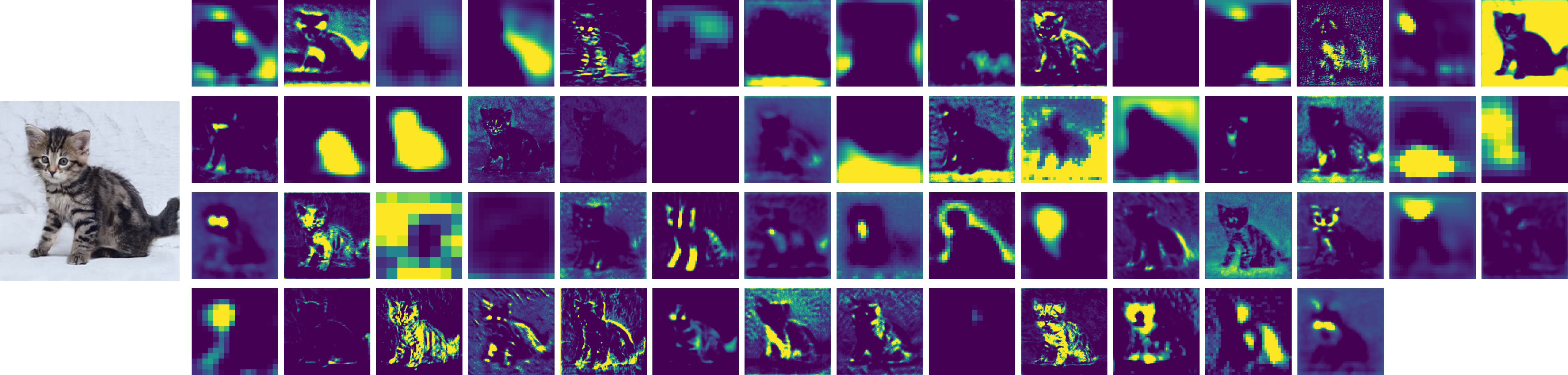}
\caption{\textbf{All the concepts for LSUN-cats.} Shown for one StyleGAN2 generated image.}
\end{figure*}
\begin{figure*}
  \centering \includegraphics[width=\textwidth]{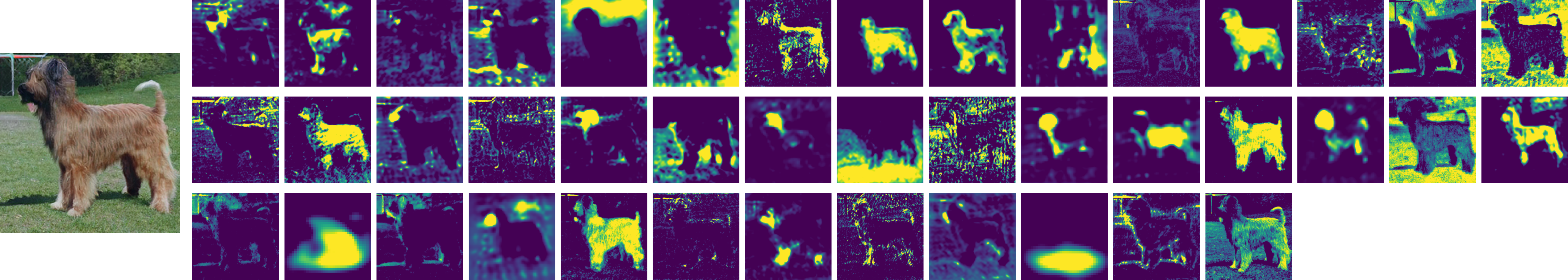}
  \label{fig:sup3}
\caption{\textbf{All the concepts for ImageNet class ``Briard''.} Shown on one StyleGAN-XL generated image.}
\end{figure*}

\begin{figure*}
  \centering \includegraphics[width=\textwidth]{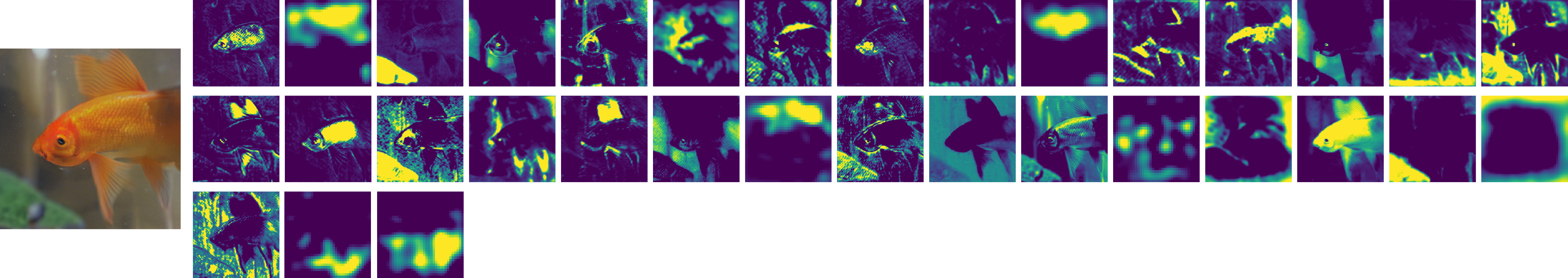}
\caption{\textbf{All the concepts for ImageNet class ``Goldfish''.} Shown on one StyleGAN-XL generated image.}
\end{figure*}
\begin{figure*}
  \centering \includegraphics[width=\textwidth]{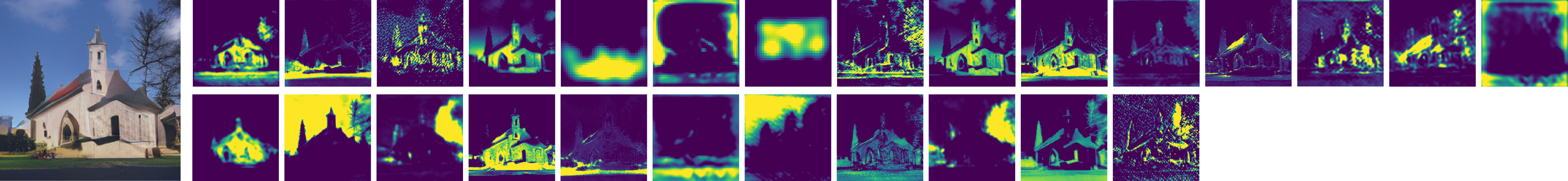}
\caption{\textbf{All the concepts for ImageNet class ``Church''.} Shown on one StyleGAN-XL generated image.}
\end{figure*}
\begin{figure*}
  \centering \includegraphics[width=\textwidth]{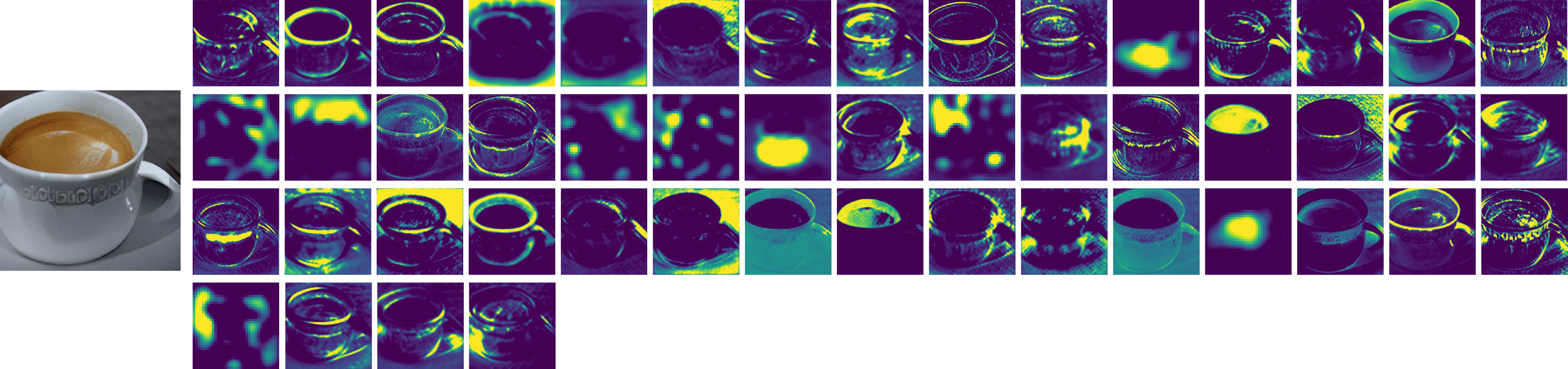}
\caption{\textbf{All the concepts for ImageNet class ``Espresso''.} Shown on one StyleGAN-XL generated image.}
\end{figure*}
\begin{figure*}
  \centering \includegraphics[width=\textwidth]{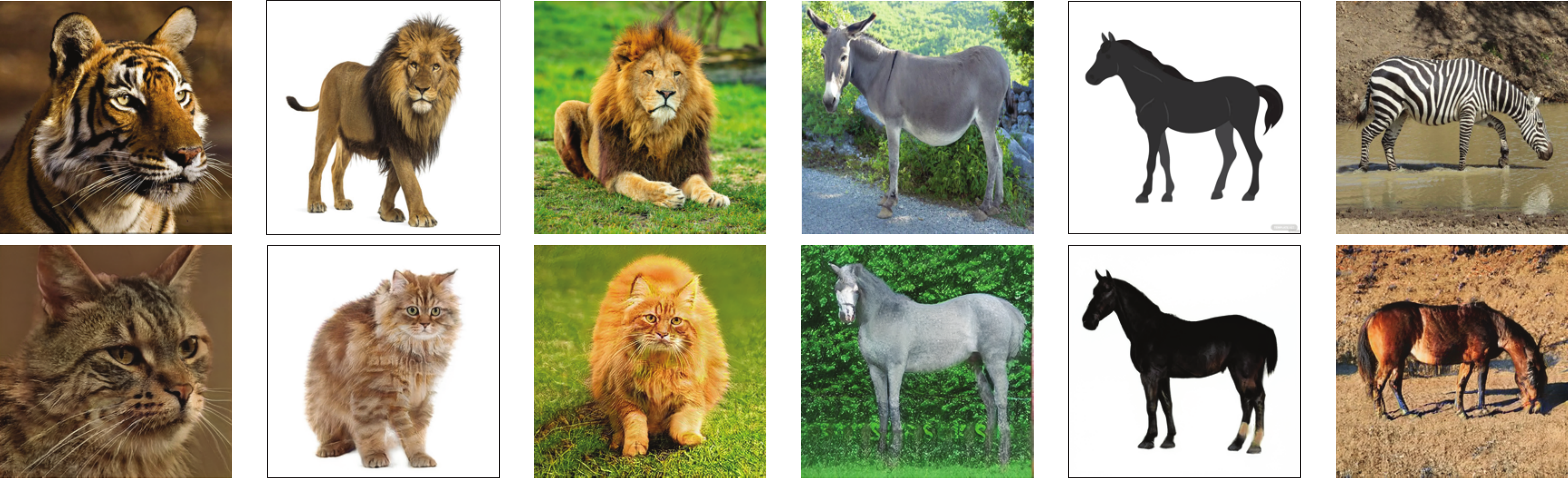}
\caption{\textbf{Additional out-of-distribution and cross-class inversions}. We show out-of-distribution image inversions done by Rosetta Neurons guidance for StyleGAN2 model, trained on LSUN cats (left 3 images) and LSUN horses (right 3 images).}
\end{figure*}
\begin{figure*}
  \centering \includegraphics[width=0.8\textwidth]{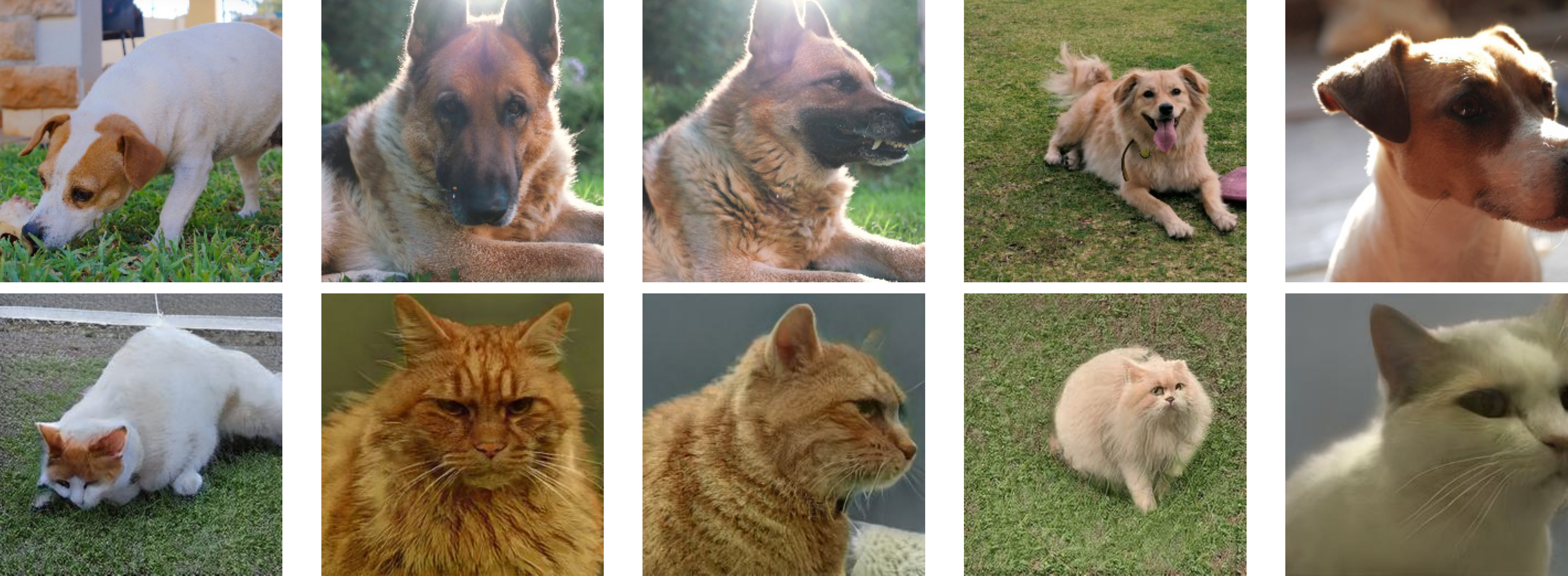}
\caption{\textbf{Dog-to-cat cross-class inversions}. Using Rosetta Neurons guidance for StyleGAN2 model, trained on LSUN cats.}
\end{figure*}
\begin{figure*}
  \centering \includegraphics[width=0.6\textwidth]{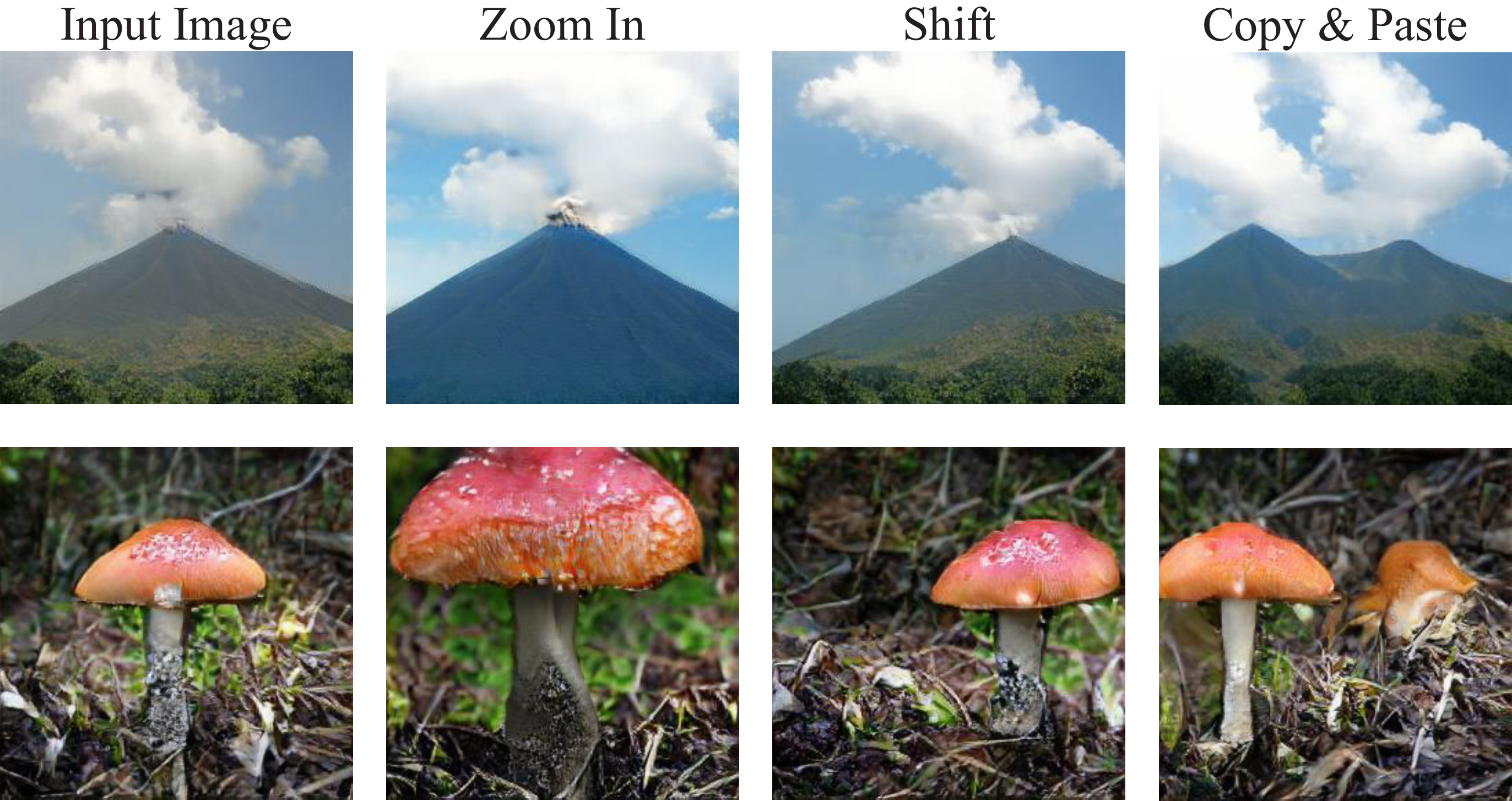}
\caption{\textbf{Additional examples of Rosetta Neurons guided editing.} We show examples using BigGAN and its matches to CLIP-RN.}
\end{figure*}
\begin{figure*}
  \centering \includegraphics[width=0.6\textwidth]{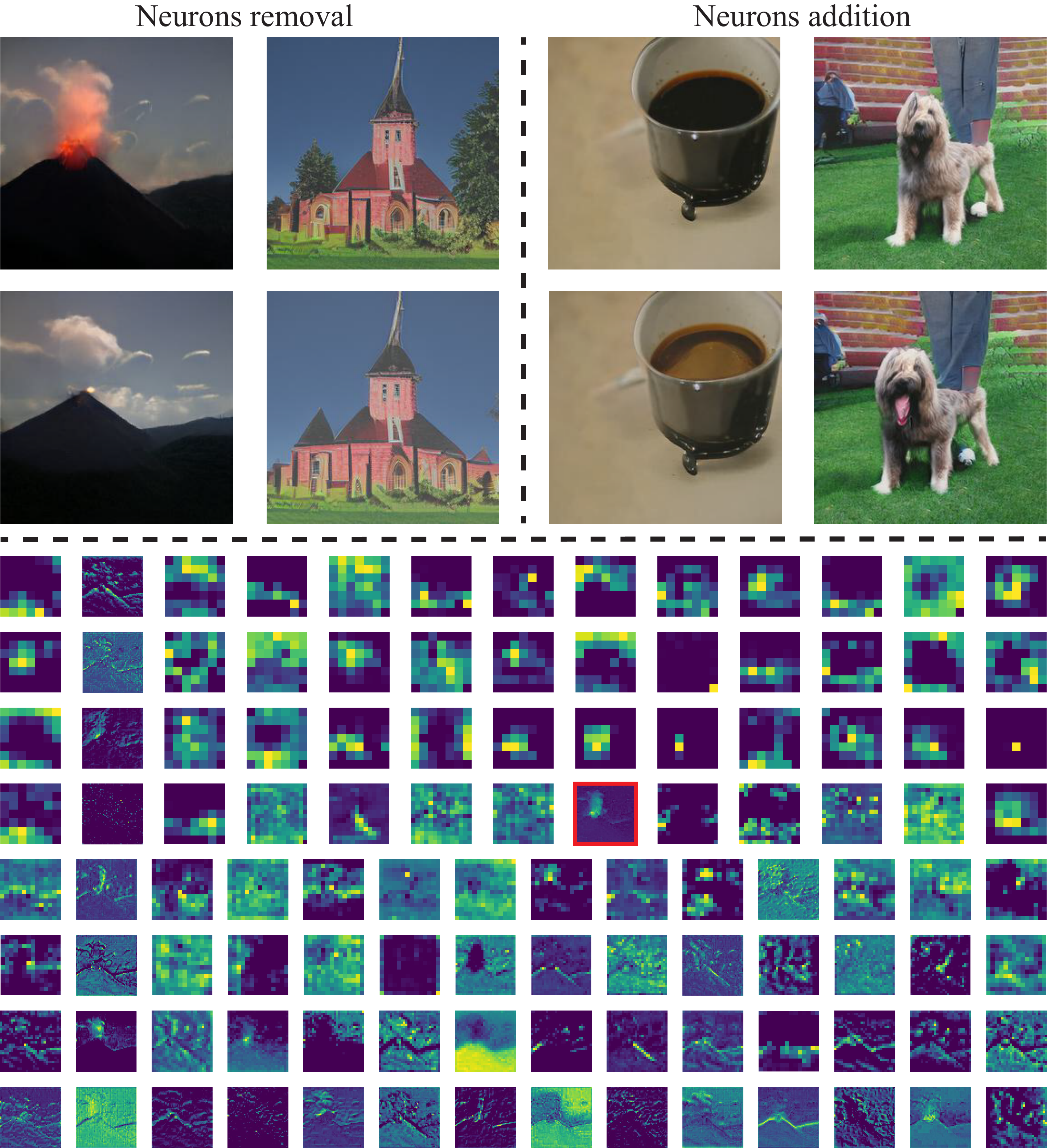}
\caption{\textbf{Additional Single Rosetta Neurons Edits}. By decreasing (two left image pairs) or increasing (two right image pairs) the values
of specific manually chosen Rosetta Neurons before the latent optimization process, we can remove or add elements to the image. In this figure, we demonstrate (left to right): Removing lava eruptions, removing trees, adding Crema to an Espresso, and adding a dog's tongue. For the leftmost example, we also provide the complete list of Rosetta Neurons visualizations. The chosen concept is marked with a red frame.}
\end{figure*}
\begin{figure*}
  \centering \includegraphics[width=0.8\textwidth]{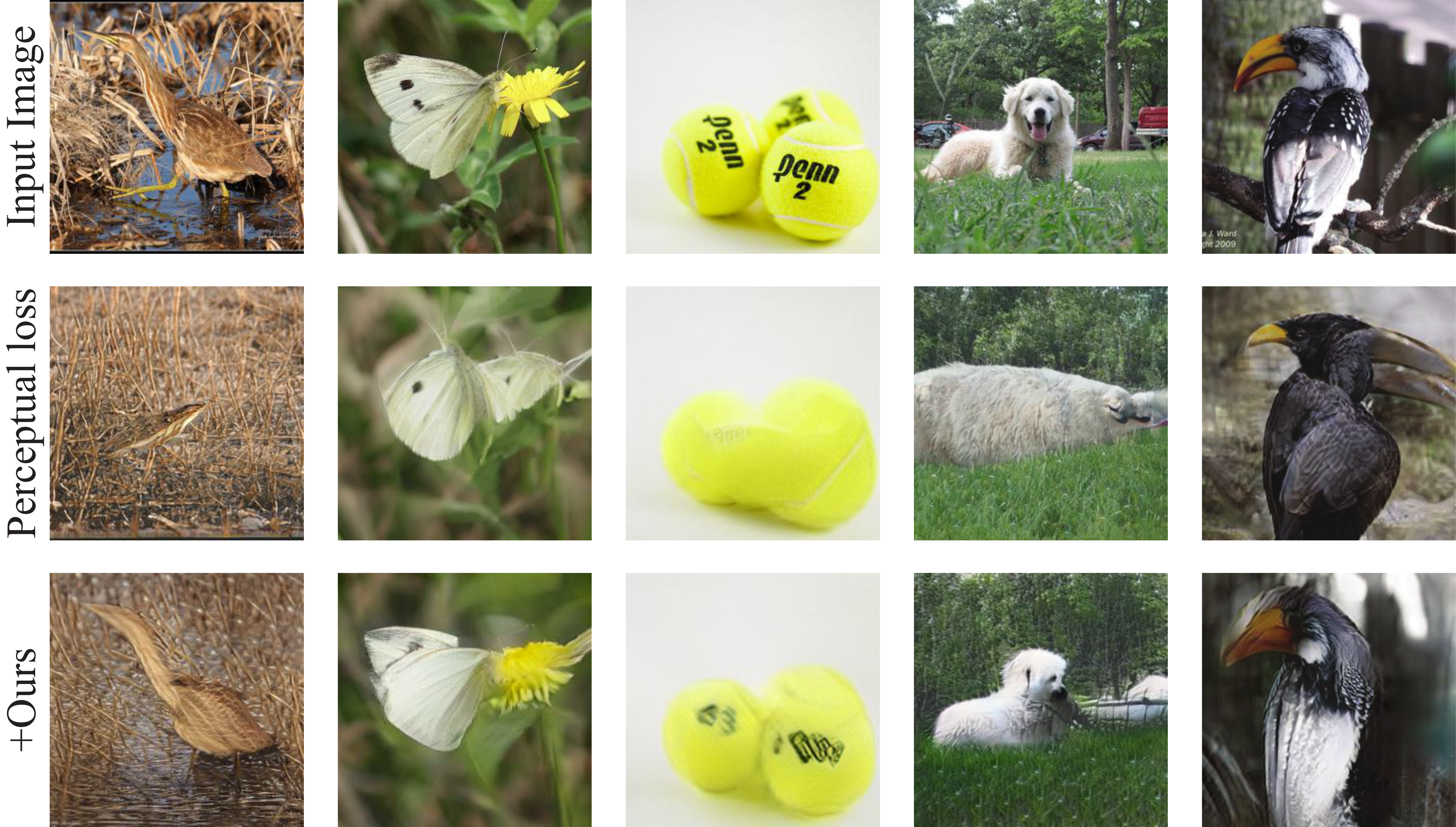}
\caption{\textbf{Additional image inversions for StyleGAN-XL.} We compare using perceptual loss (second row) to perceptual loss with additional guidance from the Rosetta Neurons (third row).}
\end{figure*}

\begin{figure*}
  \centering \includegraphics[width=0.8\textwidth]{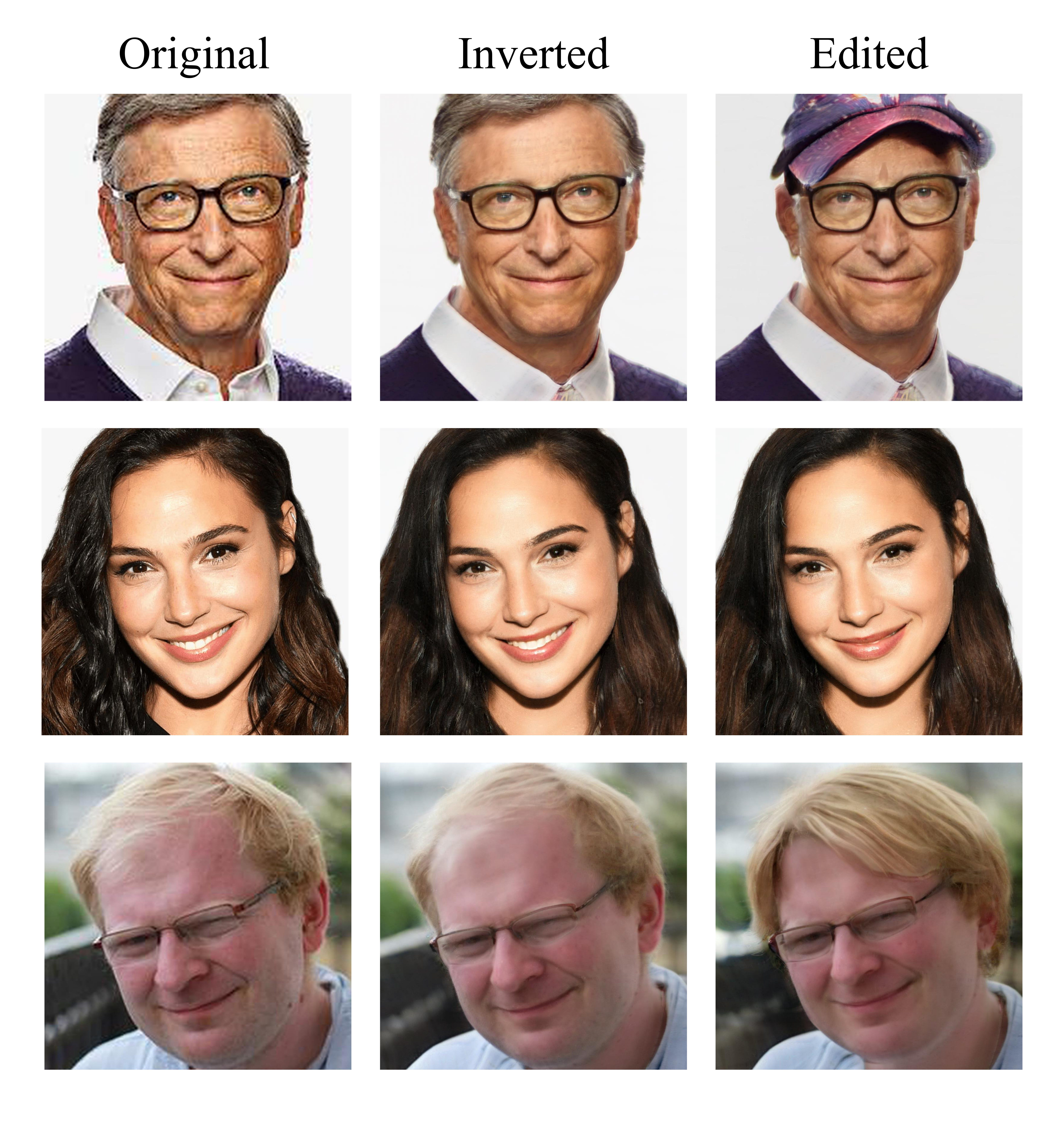}
\caption{\textbf{High Resolution single Rosetta Neuron Edits} We provide additional examples, complementary to Fig.~\ref{fig:removals}, but with higher resolution. We conduct matching between a StyleGAN3 trained on $1024$$\times$$1024$ FFHQ images and DINO-ViT with 1000 images, which takes $~2700s$.
We then apply standard PTI~\cite{roich2021pivotal} to a real high-res ($1024$$\times$$1024$) image (160s). Finally, we perform our editing which takes 18.4s (Zoom-in possible).}
\end{figure*}

\end{document}